\documentclass[conference]{IEEEtran}
\IEEEoverridecommandlockouts
\usepackage{float}
\usepackage{cite}
\usepackage{amsmath,amssymb,amsfonts}
\usepackage{algorithmic}
\usepackage{graphicx}
\usepackage{textcomp}
\usepackage{xcolor}
\usepackage{colortbl}
\usepackage{comment}
\usepackage{hyperref}
\usepackage{multirow}
\usepackage{tabularx}
\usepackage{booktabs}
\newcolumntype{C}{>{\centering\arraybackslash}X}

\def\BibTeX{{\rm B\kern-.05em{\sc i\kern-.025em b}\kern-.08em
    T\kern-.1667em\lower.7ex\hbox{E}\kern-.125emX}}
\begin{document}

\title{StereoGeo: an end-to-end stereo camera calibration method\\
}

\author{\IEEEauthorblockN{Imane MEDDOUR}
\IEEEauthorblockA{\textit{Universite Paris-Saclay,} \\
\textit{CEA, List,}\\
F-91120, Palaiseau, France \\
imane.meddour@cea.fr}
\and
\IEEEauthorblockN{Andréa MACARIO BARROS}
\IEEEauthorblockA{\textit{Universite Paris-Saclay,} \\
\textit{CEA, List,}\\
F-91120, Palaiseau, France \\
andrea.barros@cea.fr}
\and
\IEEEauthorblockN{Cédric GOUY-PAILLER}
\IEEEauthorblockA{\textit{Universite Paris-Saclay,} \\
\textit{CEA, List,}\\
F-91120, Palaiseau, France \\
cedric.gouy-pailler@cea.fr}
}

\maketitle

\begin{abstract}
In this work, we propose StereoGeo, an end-to-end network-based approach for stereo camera calibration. Our method estimates the focal lengths and gravity directions of the left and right cameras, as well as the relative extrinsic transformation relating them. Existing methods often rely on calibration patterns in structured environments or address only a single camera configuration, being limited to either intrinsic or extrinsic estimation, and depending on a multi-view setups. StereoGeo extends the GeoCalib algorithm, integrating deep neural network feature extraction with a differentiable optimizer. Extensive experiments on real-world benchmarks demonstrate that StereoGeo achieves competitive performance for intrinsic calibration and provides accurate stereo extrinsic estimation, outperforming existing methods that are limited to monocular settings. The dataset used in this work is partially publicly available at \url{https://github.com/meddourimane/StereoGeo-dataset}.
\end{abstract}

\begin{IEEEkeywords}
Stereo camera calibration, Deep learning, End-to-end learning.
\end{IEEEkeywords}

\section{Introduction}
\label{sec:intro}
Stereo camera calibration is the task of estimating the intrinsic parameters of each camera, such as focal length, principal point, and lens distortion, as well as the rigid body transformation that defines the relative pose between them. This task plays a crucial role in computer vision such as Simultaneous Localization and Mapping (SLAM), Structure from Motion (SfM), and 6D object pose tracking~\cite{liao2025deeplearningcameracalibration}. 

Stereo camera calibration pipelines can be divided into geometric and learning-based categories. 
Geometric methods often use known calibration patterns such as checkerboards or planar grids~\cite{bouguet_calib}. Among them, one of the most employed algorithms is the Zhang's method~\cite{888718}, which estimates intrinsic and extrinsic parameters from multiple planar pattern observations, providing high accuracy but requiring carefully captured images and being sensitive to noise or imperfect pattern detection.

Recent learning-based calibration approaches overcome these limitations by leveraging neural networks to predict camera parameters~\cite{Lopez_2019_CVPR, Song_2024_WACV, jin2023perspectivefieldssingleimage}. In particular, GeoCalib~\cite{veicht2024geocalib} represents a hybrid learning–geometry framework, achieving state-of-the-art performance in single-view calibration by estimating intrinsic parameters and gravity direction from a single RGB image, and refining parameters through differentiable optimization without calibration patterns or multi-view constraints.
\begin{figure}[t]
  \centering
  \includegraphics[width=1.0\linewidth]{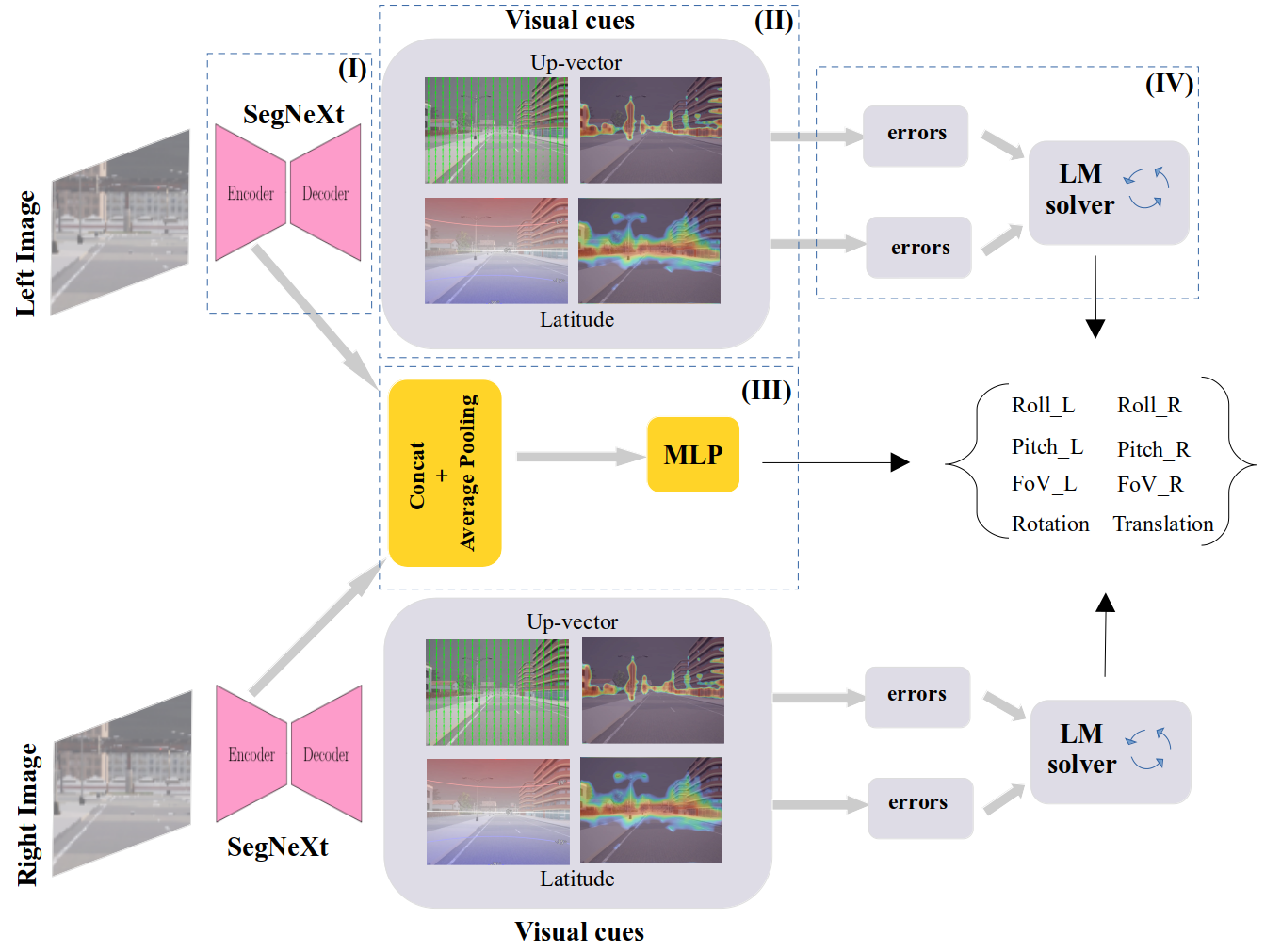}
  \caption{The StereoGeo architecture. The network predicts per-view perspective fields and camera parameters (focal, gravity, rotation and translation), which are refined using a differentiable Levenberg-Marquardt optimization.}
  \label{fig:stereogeo_arch}
\end{figure}
Nevertheless, most learning-based methods primarily focus on single-view intrinsic calibration, including GeoCalib. UGCL approach~\cite{waleed2024cameracalibrationgeometricconstraints} represents a notable effort toward stereo learning-based calibration, jointly estimating intrinsic and extrinsic parameters from stereo pairs under geometric constraint losses. However, UGCL assumes both cameras share identical intrinsic parameters, which does not hold in many practical setups where left and right cameras may have different focal lengths or principal points. Furthermore, it does not explicitly estimate per-camera gravity directions.

To address these limitations, we propose StereoGeo, a novel stereo calibration framework that extends GeoCalib to stereo configurations. StereoGeo jointly predicts per-camera focal lengths, per-camera gravity directions, and extrinsic parameters using a hybrid learning and differentiable optimization pipeline, without relying on calibration patterns or feature matching. Our method leverages deep networks to predict per-view perspective fields, which are then refined through a differentiable Levenberg–Marquardt optimization, ensuring robustness even for cameras with distinct intrinsics. Additionally, we construct a large-scale synthetic stereo calibration dataset composed of both indoor and outdoor scenes. The dataset combines stereo panorama-based image generation and CARLA-rendered images, resulting in over 50k stereo pairs with diverse geometric layouts and environmental conditions.

The paper is structured as follows: Section~\ref{sec:prob} formalizes the problem; Section~\ref{sec:stereo} describes the proposed method; Section~\ref{sec:exp} presents experiments and discusses the results; Section~\ref{sec:conc} concludes the work.

\section{Problem Formulation}
\label{sec:prob}
Accurate camera calibration is crucial for 3D perception tasks, ensuring image measurements correspond to real-world geometry. We formulate the calibration problem by modeling the camera as a sensor transforming 3D scene points into 2D image observations.

\subsection{Static Camera Sensor Model}
Let $\mathbf{X}_k \in \mathbb{R}^3$ be a 3D point in the camera coordinate system 
and $\mathbf{y}_k \in \mathbb{R}^2$ its pixel observation. The image formation 
can be expressed as:
\begin{equation}
\mathbf{y}_k = f(\mathbf{X}_k, \boldsymbol{\theta}) + \boldsymbol{\nu}_k,
\end{equation}
where $f(\cdot, \boldsymbol{\theta})$ is the non-linear projection model, $\boldsymbol{\theta}$ represents the intrinsic parameters (focal length $f_x, f_y$, 
gravity direction) and the term $\boldsymbol{\nu}_k \sim \mathcal{N}(0, \sigma^2 \mathbf{I})$ represents the additive measurement noise in the image plane, usually modeled as Gaussian white noise.
\subsection{Monocular Camera Calibration}
The goal of monocular calibration is to estimate the optimal parameters $\hat{\boldsymbol{\theta}}$ from a set of $N$ known 3D points $\mathbf{X}_k$ and their detected image projections $\mathbf{y}_k$. 
Therefore, instead of minimizing the error in the 3D scene, we minimize the reprojection error in the image plane~\cite{888718}:
\begin{equation}
\hat{\boldsymbol{\theta}} = \arg\min_{\boldsymbol{\theta}} \frac{1}{N} \sum_{k=1}^{N} \left\| \mathbf{y}_k - f(\mathbf{X}_k, \boldsymbol{\theta}) \right\|^2,
\end{equation}
where $k = 1, \ldots, N$ indexes the point correspondences. 
The noise $\boldsymbol{\nu}_k$ is implicitly handled by this least-squares formulation, which yields the Maximum Likelihood Estimate (MLE) under Gaussian assumptions.

\subsection{Stereo Camera Calibration}
In a stereo setup, two cameras, denoted as left ($L$) and right ($R$), observe the same scene points from different viewpoints. Each camera follows its own projection model:
\begin{equation}
\begin{gathered}
\mathbf{y}_{k,L} = f_L(\mathbf{X}_{k,L}, \boldsymbol{\theta}_L) + \boldsymbol{\nu}_L, \\[2pt]
\mathbf{y}_{k,R} = f_R(\mathbf{X}_{k,R}, \boldsymbol{\theta}_R) + \boldsymbol{\nu}_R.
\end{gathered}
\end{equation}
The 3D coordinates of a point in the two camera frames are related by a rigid body transformation:
\begin{equation}
\mathbf{X}_{k,R} = R_s \mathbf{X}_{k,L} + \mathbf{t}_s,
\end{equation}
where $R_s \in SO(3)$ (the Special Orthogonal group of 3×3 rotation matrices) is the relative rotation between the cameras, 
$\mathbf{t}_s \in \mathbb{R}^3$ is the relative translation vector.\\
The calibration problem for a stereo system consists of jointly estimating the intrinsics $(\boldsymbol{\theta}_L, \boldsymbol{\theta}_R)$ and the relative transformation $(R_s, \mathbf{t}_s)$ by minimizing the total reprojection error:
\begin{equation}
\begin{aligned}
\min_{\substack{
\boldsymbol{\theta}_L, \boldsymbol{\theta}_R,\\
R_s, \mathbf{t}_s
}}
\sum_{k=1}^{N} 
\Big(
&\left\| \mathbf{y}_{k,L} - f_L(\mathbf{X}_{k,L}, \boldsymbol{\theta}_L) \right\|^2 \\
&+ \left\| \mathbf{y}_{k,R} - f_R(R_s \mathbf{X}_{k,L} + \mathbf{t}_s, \boldsymbol{\theta}_R) \right\|^2
\Big).
\end{aligned}
\end{equation}
\section{Stereo Calibration Framework}
\label{sec:stereo}
\subsection{StereoGeo Architecture}
StereoGeo addresses joint intrinsic and extrinsic calibration from stereo image pairs without relying on calibration patterns or explicit feature correspondences.
Given a pair of uncalibrated images $(\mathbf{I}_L, \mathbf{I}_R)$ acquired by stereo camera, the proposed framework estimates for each camera the intrinsic parameters (focal length), gravity directions $(\mathbf{g}_L, \mathbf{g}_R)$ decomposed into roll and pitch, and the relative pose $(\mathbf{R}_s, \mathbf{t}_s)$ between the cameras.

As illustrated in Fig.~\ref{fig:stereogeo_arch}, the proposed architecture is organized into four main stages: (I) Independent camera-wise feature extraction,  
(II) Perspective field, (III) Stereo-view fusion for relative pose estimation, and (IV) Optimization via Levenberg-Marquardt (LM). Following, each stage is detailed:\\
\\
\textbf{(I) Independent camera-wise feature extraction: }The left and right images are processed independently by two identical encoder–decoder networks.
This design maintains the modularity of each camera branch and avoids any cross-view information exchange during the estimation of intrinsic parameters and gravity directions. We employ a SegNeXt~\cite{guo2022segnextrethinkingconvolutionalattention} based encoder–decoder architecture, adapted to preserve the original spatial resolution of the input images via progressive upsampling and low-level feature fusion, which is critical for dense geometric prediction. We chose SegNeXt for its multi-scale convolutional attention mechanism, which effectively captures both local vanishing point cues and global scene geometry essential for gravity estimation.\\
\textbf{(II) Perspective Fields: }Perspective Fields~\cite{jin2023perspectivefieldssingleimage, veicht2024geocalib} provide a dense geometric representation of the scene by encoding, for each pixel, the local gravity direction and viewing angle with respect to the world coordinate system. Following this formulation, each camera branch based on SegNeXt~\cite{guo2022segnextrethinkingconvolutionalattention} predicts up-vectors $\hat{\mathbf{u}}_p$ and latitude values $\hat{\phi}_p$ for each pixel $p$. Each pixel $p \in \mathbb{R}^2$ on the image frame corresponds to a light ray $\mathbf{n} \in \mathbb{R}^3$ emitted from a 3D point $\mathbf{X} \in \mathbb{R}^3$ in the world frame. The up-vector represents the projection of the gravity direction into the image plane and can be inferred from both low-level geometric structures (e.g., vertical edges, line segments) and high-level semantic cues (e.g., upright objects). The latitude encodes the angle between the viewing ray and the horizontal plane, with zero latitude corresponding to the horizon. Formally, considering a pixel $p$ observing a 3D point $\mathbf{X} \in \mathbb{R}^3$ under gravity direction
$\mathbf{g}$, these quantities are defined as showed by Eqs.~\eqref{eq:upvector} and  \eqref{eq:phi}~\cite{jin2023perspectivefieldssingleimage}:
\begin{align}
\mathbf{u}_p &=
\lim_{c \to 0}
\frac{\pi(\mathbf{X} - c \mathbf{g}) - \pi(\mathbf{X})}
{\left\| \pi(\mathbf{X} - c \mathbf{g}) - \pi(\mathbf{X}) \right\|_2},
\label{eq:upvector}
\\
\phi_p &=
\arcsin\!\left(
\frac{\mathbf{n}^\top \mathbf{g}}{\|\mathbf{n}\|_2}
\right).
\label{eq:phi}
\end{align}
where $\pi(\cdot)$ denotes the camera projection function and $\mathbf{n}$ is the corresponding viewing ray.\\
In addition to these fields, the network predicts pixel-wise confidence maps $(\sigma_{\mathbf{u}_p}, \sigma_{\phi_p})$ that quantify the reliability of each prediction.
These confidences allow subsequent optimization to focus on informative regions (e.g., near vertical structures or the horizon) while down-weighting ambiguous or textureless areas.\\
\textbf{(III) Stereo-view fusion for relative pose estimation: }To estimate the extrinsic parameters, StereoGeo employs a dedicated stereo-view fusion module that correlates features from the left and right encoder branches. Our design preserves the independence of monocular intrinsic and gravity predictions, using stereo-view information exclusively for pose estimation.\\
Let $\mathbf{F}_L$ and $\mathbf{F}_R$ denote the feature maps from the left and right encoders. These features are concatenated along the channel dimension and aggregated through global pooling. The resulting feature vector is then passed to two separate MLP-based regression heads for rotation and translation prediction.\\
Rotation is predicted as a unit quaternion $\mathbf{q}_s = [q_x, q_y, q_z, q_w]^T$ and normalized via $L_2$ normalization to ensure a valid representation on the unit sphere. This quaternion is subsequently converted to the rotation matrix $R_s \in SO(3)$. Translation is predicted as a 3D vector $\mathbf{t}_s = [t_x, t_y, t_z]^\top$ that encodes both magnitude and direction.\\
\textbf{(IV) Optimization via Levenberg-Marquardt: }Given the predicted Perspective Fields $(\hat{\mathbf{u}}_p, \hat{\phi}_p)$ from the SegNeXt encoder--decoder, we refine the camera parameters $\boldsymbol{\theta} = \{f, \mathbf{g}\}$ (focal length and gravity direction) for each camera through non-linear optimization. Since each camera has independent parameters, we define separate objective functions for the left and right views.\\
For a given camera, let $\mathbf{u}_p(\theta)$ and $\phi_p(\theta)$ denote the Perspective Fields induced by the current parameters. We define the confidence-weighted per-pixel residuals as: 
\begin{equation}
\mathbf{r}_{\mathbf{u}_p} = \mathbf{u}_p(\theta) - \hat{\mathbf{u}}_p, \quad
r_{\phi_p} = \sin \phi_p(\theta) - \sin \hat{\phi}_p,
\end{equation}
and minimize the per-camera objective function:
\begin{equation}
E_{\text{cam}}(\theta) =
\sum_{p \in H \times W} \sigma_{\mathbf{u}_p} \|\mathbf{r}_{\mathbf{u}_p}\|_2^2 + \sigma_{\phi_p} \|r_{\phi_p}\|_2^2,
\end{equation}
where $\sigma_{\mathbf{u}_p}$ and $\sigma_{\phi_p}$ are the predicted confidence maps.\\
We employ the Levenberg--Marquardt (LM) algorithm~\cite{Levenberg1944AMF,doi:10.1137/0111030} to minimize $E_{\text{cam}}$. The gravity vector $\mathbf{g}$ is parametrized on the unit sphere $\mathbb{S}^2$~\cite{Hartley_Zisserman_2004, HERTZBERG201357}, and the focal length as $\log f$ to enforce positivity. At each iteration, the LM update is computed as:
\begin{equation}
\delta = - (\mathbf{H} + \lambda \operatorname{diag}(\mathbf{H}))^{-1} \mathbf{J}^\top \mathbf{W} \mathbf{r},
\end{equation}
where $\mathbf{r}$ stacks all residuals, $\mathbf{W}$ is the diagonal confidence weight matrix, $\mathbf{J}$ is the Jacobian, $\mathbf{H} = \mathbf{J}^\top \mathbf{W} \mathbf{J}$ is the Hessian, and $\lambda$ is the damping factor. Following~\cite{veicht2024geocalib}, we initialize gravity to $\mathbf{g}_0 = [0, 1, 0]^\top$ and focal length to $f_0 = 0.7 \cdot \max(W, H)$. The optimization terminates when the update $\delta$ becomes sufficiently small, producing refined parameters $\theta_L$ and $\theta_R$ for the left and right cameras.
\subsection{Loss Formulation and Training Strategy}
StereoGeo is trained in a fully supervised, end-to-end manner. Thanks to the differentiability of the Levenberg--Marquardt (LM), gradients can be backpropagated through the refinement layer, enabling joint learning of Perspective Fields and camera parameters.
\subsubsection{Supervision of intrinsic parameters and Perspective Fields}
For each camera, let $(\hat{\mathbf{u}}_p, \hat{\phi}_p)$ denote the predicted Perspective Fields and $\hat{\theta}$ the refined parameters (focal length and gravity) after LM optimization. Given ground-truth parameters $\bar{\theta}$, we define a per-camera loss that jointly supervises the refined parameters and intermediate geometric cues, as shown in Eq.~\eqref{eq:camera_loss}:
\begin{align}
\mathcal{L}_{\text{cam}} &= 
\|\hat{\theta} - \bar{\theta}\|
+ \beta \sum_{p \in H \times W} \sigma_{\mathbf{u}_p} \|\hat{\mathbf{u}}_p - \mathbf{u}(\bar{\theta})_p\| \notag\\
&\quad + \sigma_{\phi_p} \|\hat{\phi}_p - \phi(\bar{\theta})_p\|,
\label{eq:camera_loss}
\end{align}
where $\beta$ balances the Perspective Field contribution and $\sigma_{\mathbf{u}_p}, \sigma_{\phi_p}$ are the predicted confidence maps.
\subsubsection{Supervision of stereo relative pose}
\begin{table*}[t]
\centering
\caption{Per-camera intrinsic calibration performance on the test set.}
\small
\setlength{\tabcolsep}{3pt}
\renewcommand{\arraystretch}{1.0}
\begin{tabularx}{\textwidth}{c l CCCCCCCCCCCC}
\hline
 & \multirow{2}{*}{Approach} 
 & \multicolumn{4}{c}{Roll [degrees]} 
 & \multicolumn{4}{c}{Pitch [degrees]} 
 & \multicolumn{4}{c}{FoV [degrees]} \\
%\cline{3-6} \cline{7-10} \cline{11-14}
\cmidrule(lr){3-6} \cmidrule(lr){7-10} \cmidrule(lr){11-14}

 &  
 & err ↓ & AUC@1 & @5 & @10 ↑
 & err ↓ & AUC@1 & @5 & @10 ↑
 & err ↓ & AUC@1 & @5 & @10 ↑ \\
\hline

% ===================== Jeu de Test =====================
& StereoGeo Right (ours) & 0.56& 70.5 &84.8 & 90.2 & 0.75 & 57.0 & 76.6 & 85.5 & 1.82& 31.7 &58.8  &75.7  \\
& StereoGeo Left (ours) & 0.48 &75.3 &87.2  &91.4  &0.67  & 60.0 & 78.7 & 86.8 & \textbf{1.57} & \textbf{35.9} & \textbf{62.8 } & \textbf{78.2}  \\
& GeoCalib~\cite{veicht2024geocalib} & \textbf{0.45} & \textbf{77.2 } & \textbf{88.4} & \textbf{92.2} & \textbf{0.63} & \textbf{62.4} & \textbf{79.6} & \textbf{87.3} & 1.65 & 33.8 & 61.4 & 77.4\\
\hline
\end{tabularx}
\label{tab:comparison}
\end{table*}
We supervise the predicted rotation $\hat{R}_s$ and translation $\hat{\mathbf{t}}_s$ with respect to ground-truth $(\bar{R}_s, \bar{\mathbf{t}}_s)$ using Huber loss $\mathcal{H}(\cdot)$ to improve robustness, as proposed in~\cite{yin2024srposetwoviewrelativepose}. For rotation, we compute the error in the rotation angle as defined in Eq.~\eqref{eq:rot}:
\begin{equation}
\mathcal{L}_{R_s} = \mathcal{H}\left(\arccos \left( \frac{\mathrm{Tr}({\hat{R}^\top}_s \bar{R}_s) - 1}{2}\right)\right),
\label{eq:rot}
\end{equation}
For translation, the error is computed in both normalized and unnormalized forms. To further enhance accuracy, we also incorporate the angular error of translation, resulting in the following three loss terms~\cite{yin2024srposetwoviewrelativepose} described in Eq.~\eqref{eq:trans}: 
\begin{equation}
\begin{aligned}
\mathcal{L}_{\text{trans}} 
&= \lambda_t \mathcal{H}(\hat{\mathbf{t}}_s - \bar{\mathbf{t}}_s) \\
&\quad + \lambda_{t_n} \mathcal{H}\left(
\frac{\hat{\mathbf{t}}_s}{\|\hat{\mathbf{t}}_s\|}
- \frac{\bar{\mathbf{t}}_s}{\|\bar{\mathbf{t}}_s\|}
\right) \\
&\quad + \lambda_{t_a} \mathcal{H}\left(
\arccos \frac{\hat{\mathbf{t}}_s \cdot \bar{\mathbf{t}}_s}
{\|\hat{\mathbf{t}}_s\| \|\bar{\mathbf{t}}_s\|}
\right).
\end{aligned}
\label{eq:trans}
\end{equation}
where $\lambda_t$, $\lambda_{t_n}$, and $\lambda_{t_a}$ are scalar coefficients used to balance the contributions of the different loss terms. The total pose loss is $\mathcal{L}_{\text{pose}} = \mathcal{L}_{R_s} + \mathcal{L}_{\text{trans}}$.
\subsubsection{Full training loss}
The final loss combines the per-camera losses for left and right views with the relative pose loss:
\begin{equation}
\mathcal{L} = \mathcal{L}_{\text{cam}}^L + \mathcal{L}_{\text{cam}}^R + \mathcal{L}_{\text{pose}}.
\end{equation}
\section{Experiments}
\label{sec:exp}
We evaluate StereoGeo on joint intrinsic and extrinsic calibration from stereo pairs across diverse scenarios. First, we assess intrinsic parameter and gravity direction estimation in the monocular case. Second, we evaluate joint intrinsic and extrinsic parameter recovery in the stereo setting. Finally, we analyze extrinsic parameter estimation independently, comparing against established geometric baselines.
\subsection{Implementation Details}
\textbf{Training Dataset: }Following~\cite{veicht2024geocalib, jin2023perspectivefieldssingleimage}, we construct a large-scale synthetic stereo calibration dataset combining stereo panorama-based rendering and photorealistic simulation. The dataset contains both indoor and outdoor environments to ensure strong geometric diversity and robustness across scene types. From 2,500 stereo equirectangular panoramas collected from SUNCG~\cite{song2016semanticscenecompletionsingle} via 3D60~\cite{zioulis2019spherical}, we generate 38,324 stereo pairs by sampling 16 crops per panorama. For each crop, roll and pitch are uniformly sampled within $[\pm 45^\circ]$, and the vertical field of view within $[20^\circ, 105^\circ]$. The stereo baseline is fixed around $0.26$~m. Since some panoramas contain padding artifacts, stereo pairs with more than $1 \%$ black pixels are discarded.
To complement panorama-based data with realistic driving scenarios, we generated 12,220 stereo pairs using CARLA~\cite{Dosovitskiy17}. Scenes are rendered across multiple towns and weather conditions, providing diverse outdoor urban layouts. For each capture, a physically consistent stereo rig with variable baseline is instantiated. The baseline is uniformly sampled in $[0.20, 0.70]$~m, the base orientation of the stereo rig is defined by roll and pitch angles uniformly sampled in $[\pm 0.5^\circ]$, and the vertical field of view is sampled in $[55^\circ, 75^\circ]$.
Additionally, 5,000 stereo pairs are extracted from TartanAir~\cite{wang2020tartanairdatasetpushlimits} across 12 diverse simulated environments, further increasing variability in scene structure and viewpoint distribution. In total, the dataset contains 55,913 stereo pairs spanning both indoor and outdoor environments. The data are split into $90\%$ training, $5\%$ validation, and $5\%$ test sets.

\textbf{Training: }StereoGeo is trained end-to-end from scratch without pre-training or fine-tuning from GeoCalib for 48 epochs using the AdamW~\cite{loshchilov2019decoupledweightdecayregularization} optimizer with a learning rate of $10^{-4}$ and a batch size of 24 stereo pairs. The learning rate is linearly warmed up over the first 4{,}000 steps and decayed by a factor of 10 at steps 80{,}000 and 130{,}000. Training requires approximately 48 hours on two NVIDIA H200 GPUs. Inference takes about 464~ms per stereo pair.
\subsection{Intrinsic Parameter and Gravity Direction Estimation (Monocular Case)}
We evaluate per-camera intrinsic calibration by comparing StereoGeo with the single-view SOTA method GeoCalib~\cite{veicht2024geocalib}, retrained on our data. We report median angular errors and Area Under the Curve (AUC) at $1^\circ$, $5^\circ$, and $10^\circ$ thresholds for roll, pitch, and vertical field-of-view (vFoV). Left and right cameras are evaluated independently on the held-out test set.
Table~\ref{tab:comparison} shows that StereoGeo achieves median errors of $0.48^\circ$/$0.67^\circ$/$1.57^\circ$ for the left camera and $0.56^\circ$/$0.75^\circ$/$1.82^\circ$ for the right camera (roll/pitch/vFoV). These results confirming that StereoGeo achieves comparable SOTA single-view calibration quality while additionally estimating stereo extrinsics. The small differences between left and right cameras demonstrate the stability of the dual-branch architecture.
\subsection{Intrinsic and Extrinsic Parameter Estimation (Stereo Case)}
We evaluate StereoGeo on relative pose and intrinsic parameter recovery for 
stereo cameras using both our held-out test set and the KITTI 
dataset~\cite{6248074}. Results are compared with 
UGCL~\cite{waleed2024cameracalibrationgeometricconstraints}, a method jointly 
estimating intrinsic and extrinsic parameters and presented on Sec.~\ref{sec:intro}.

\begin{table}[t]
\centering
\caption{Stereo intrinsic and extrinsic calibration.}
\scriptsize
\setlength{\tabcolsep}{3pt}
\renewcommand{\arraystretch}{1.1}
\begin{tabular}{l l c c c c c c}
\hline
\multirow{2}{*}{Dataset} & \multirow{2}{*}{Method} & \multicolumn{2}{c}{vFoV Med. (°)} & Pitch Med. (°) & \multicolumn{3}{c}{Trans. Error (m)} \\
\cline{3-8}
 & & L & R & & Mean & Med. & Std \\
\hline
\multirow{2}{*}{Test Set} 
& UGCL & 14.68 & 14.68 & 6.39 & 0.40 & 0.36 & 0.10 \\
& StereoGeo (ours)& \textbf{1.61} & \textbf{1.82} & \textbf{0.10} & \textbf{0.024} & \textbf{0.012} & \textbf{0.039} \\
\hline
\multirow{2}{*}{KITTI} 
& UGCL & \textbf{5.04} & \textbf{5.04} & 6.35 & 0.62 & 0.62 & \textbf{0.0001} \\
& StereoGeo (ours) & 10.81 & 10.82 & \textbf{0.18} & \textbf{0.40} & \textbf{0.40} & 0.070 \\
\hline
\end{tabular}
\label{tab:extrinsic_in}
\end{table}
As shown in Table~\ref{tab:extrinsic_in}, StereoGeo recovers the full rotation matrix and per-camera focal lengths without feature matching or RANSAC~\cite{10.1145/358669.358692}. On our test\nolinebreak\ set, StereoGeo outperforms UGCL across all metrics, demonstrating its effectiveness on diverse synthetic scenarios. On KITTI, UGCL achieves lower vFoV error due to the dataset's fixed camera optics and limited focal length variability. However, StereoGeo remains significantly more accurate in estimating extrinsic parameters, which are critical for stereo vision applications. These results highlight StereoGeo as a robust learning-based method that provides superior relative pose estimation, even on real-world data with different characteristics from the training set.
\subsection{Extrinsic Parameter Estimation}
We evaluate relative pose estimation on KITTI Stereo~\cite{6248074} using 200 image pairs against feature-based geometric pipelines that use RANSAC for robust Essential Matrix estimation: ORB~\cite{6126544} and SuperPoint with 
SuperGlue~\cite{detone2018superpointselfsupervisedpointdetection,sarlin2020supergluelearningfeaturematching}. These geometric methods assume known intrinsic parameters, whereas StereoGeo jointly estimates both intrinsics and extrinsics. We report rotation errors with mean, median, and standard deviation. Translation results are presented in Table~\ref{tab:extrinsic_in}.

\begin{table}[t] 
\centering
\caption{Extrinsic parameter estimation on KITTI Stereo.}
\scriptsize
\setlength{\tabcolsep}{4pt}
\renewcommand{\arraystretch}{1.1}
\begin{tabular}{l ccc}
\hline
\multirow{2}{*}{Method}
& \multicolumn{3}{c}{Rotation Error (°)} \\
\cline{2-4}
& Mean & Med. & Std  \\
\hline
ORB + EM & \textbf{0.57} & 0.47 & 0.30  \\
SuperPoint + SuperGlue & 1.43 & \textbf{0.46} & 12.64 \\
StereoGeo (ours) & \textbf{0.57} & 0.58 & \textbf{0.094} \\
\hline
\end{tabular}
\label{tab:extrinsic}
\end{table}
As shown in Table~\ref{tab:extrinsic}, StereoGeo achieves competitive rotation accuracy with mean error matching ORB with EM and median error close to SuperPoint with SuperGlue. Notably, StereoGeo exhibits lower standard deviation ($0.094^\circ$) compared to both baselines ($0.30^\circ$ and $12.64^\circ$), demonstrating superior robustness and consistency. Critically, unlike geometric baselines that cannot recover metric translation due to scale ambiguity, StereoGeo recovers metric translation 
(Table~\ref{tab:extrinsic_in}, KITTI row). This demonstrates that StereoGeo provides a robust, learning-based alternative for complete extrinsic calibration in stereo setups without requiring feature matching, RANSAC, or prior knowledge of camera parameters.
\section{Conclusion}
\label{sec:conc}
We presented StereoGeo, the first learning-based framework for pattern-free stereo camera calibration, jointly estimating per-camera focal lengths, gravity directions, and extrinsic parameters without assuming identical intrinsics. By extending GeoCalib to the stereo setting, our dual-branch architecture inherits end-to-end differentiable geometric optimization and learned confidence weighting, while addressing the additional challenge of joint intrinsic and extrinsic estimation. StereoGeo operates without calibration patterns, feature matching, or multi-view constraints, and recovers metric translation directly from image. Experiments on synthetic benchmarks and real-world sequences validate the approach, showing robust extrinsic estimation and competitive intrinsic accuracy relative to single-view baselines. Current limitations, notably the absence of lens distortion modeling and principal point estimation, are natural directions for future extensions. Beyond stereo cameras, this work opens a path toward learning-based calibration of more complex multi-sensor systems in unconstrained environments.
\section{Acknowledgements}
\label{sec:ack}
CGP has benefited from French State aid managed by the Agence Nationale de la Recherche (ANR) under France 2030 program with the reference ANR-23-PEIA-005 (REDEEM project).
\bibliographystyle{IEEEtran}
\bibliography{references}
\end{document}